\documentclass[11pt,a4paper]{article}
\usepackage[hyperref]{acl2019}
\usepackage{times}
\usepackage{latexsym}
\usepackage{multirow}
\usepackage{amsmath}
\usepackage{rotating}
\usepackage{verbatim}
\usepackage{tabularx}

\usepackage{paralist}

\usepackage{xr}

\makeatletter
\newcommand*{\addFileDependency}[1]{
  \typeout{(#1)}
  \@addtofilelist{#1}
  \IfFileExists{#1}{}{\typeout{No file #1.}}
}
\makeatother
 
\newcommand*{\myexternaldocument}[1]{%
    \externaldocument{#1}%
    \addFileDependency{#1.tex}%
    \addFileDependency{#1.aux}%
}

\myexternaldocument{supplemental}

\usepackage{url}

\usepackage{footnote}
\usepackage{xspace}
\usepackage{tikz}
\usepackage{tikz-qtree-compat}
\usetikzlibrary{shapes, fit, arrows.meta}
\usepackage{lingmacros}
\usepackage{colortbl} 
\usepackage{tabularx}
\usepackage{flafter} 
\usepackage{float}
\usepackage{graphicx}
\usepackage[skins]{tcolorbox}
\tikzset{>=latex}
\usepackage{footmisc}

\newsavebox{\fmbox}

\usepackage[color=red!5]{todonotes}
\let\todonote\todo

\newcommand{\inlinetext}[1]{``\textit{#1}''}

\usepackage{url}

\newenvironment{examplebox}[1]{\begin{tcolorbox}[title=#1, fonttitle=\bfseries, colback=blue!5!white, colframe=blue!75!black, colbacktitle=blue!75!black]}{\end{tcolorbox}}

\aclfinalcopy 

\usepackage{xcolor}
\usepackage[skins]{tcolorbox}
\usepackage{hyperref}

\definecolor{darkblue}{rgb}{0, 0, 0.5}
\hypersetup{colorlinks=true,citecolor=darkblue, linkcolor=darkblue, urlcolor=darkblue}

\usepackage{float}
\usepackage{graphicx}
\usepackage{wrapfig}
\usepackage{floatflt}
\input{insbox.tex}

\usepackage{multirow}
\usepackage{xspace} 
\usepackage{fancyvrb} 
\usepackage{caption} 
\usepackage{xstring}
\usepackage{soul}
\usepackage[normalem]{ulem} 
\usepackage{paralist} 
\usepackage{algpseudocode,algorithm} 
\usepackage{tikz}
\usepackage{tikz-qtree-compat}
\usetikzlibrary{shapes, fit, arrows.meta}
\usepackage{lingmacros}
\usepackage{colortbl} 
\usepackage{tabularx}
\usepackage{flafter} 

\usepackage{adjustbox}

\algnewcommand{\LineComment}[1]{\State \(\triangleright\) #1} 

\algrenewcommand\alglinenumber[1]{\scriptsize #1:} 

\tikzset{>=latex}

\renewcommand{\todo}[1]{\todonote[size=\tiny]{#1}{\textcolor{red}{(TODO: #1)}}}
\newcommand{\annot}[2][]{\todonote[size=\tiny]{#1}{\textcolor{red}{(#2)}}}

\usepackage{listings}
\usepackage{fancyvrb}

\title{Transforming Complex Sentences into a Semantic Hierarchy}

\author{Christina Niklaus\textsuperscript{1}\textsuperscript{3}, Matthias Cetto\textsuperscript{1}, Andr\'{e} Freitas\textsuperscript{2}, \and Siegfried Handschuh\textsuperscript{1}\textsuperscript{3} \\
  \textsuperscript{1} University of St.Gallen\\
  {\small{{\tt \{christina.niklaus, matthias.cetto, siegfried.handschuh\}{\tt @unisg.ch}}}}\\
  \textsuperscript{2} University of Manchester\\
  {\small{{\tt andre.freitas@manchester.ac.uk}}}\\
  \textsuperscript{3} University of Passau\\
  {\small{{\tt \{christina.niklaus, siegfried.handschuh\}{\tt @uni-passau.de}}}}
\\}

\date{}

\begin{document}
\maketitle
\begin{abstract}
We present an approach for recursively splitting and rephrasing complex English sentences into a novel semantic hierarchy of simplified sentences, with each of them presenting a more regular structure that may facilitate a wide variety of artificial intelligence tasks, such as machine translation (MT) or information extraction (IE). Using a set of hand-crafted transformation rules, input sentences are recursively transformed into a two-layered hierarchical representation in the form of core sentences and accompanying contexts that are linked via rhetorical relations. In this way, the semantic relationship of the decomposed constituents is preserved in the output, maintaining its interpretability for downstream applications. Both a thorough manual analysis and automatic evaluation across three datasets from two different domains demonstrate that the proposed syntactic simplification approach outperforms the state of the art in structural text simplification. 
Moreover, an extrinsic evaluation shows that when applying our framework as a preprocessing step the performance of state-of-the-art Open IE systems can be improved by up to 346\% in precision and 52\% in recall. To enable reproducible research, all code is provided online.
\end{abstract}

\section{Introduction}


Text Simplification (TS) is defined as the process of reducing the linguistic complexity  of natural language (NL) text by utilizing a more readily accessible vocabulary and sentence structure. 
Its goal is to improve the readability of a text, making information easier to comprehend for people with reduced literacy, such as non-native speakers \cite{Paetzold:2016:ULS:3016387.3016433}, aphasics \cite{carroll1998practical}, dyslexics \cite{rello2013impact} or deaf persons \cite{Inui:2003:TSR:1118984.1118986}.
However, not only human readers may benefit from TS. Previous work has established that applying TS as a preprocessing step can improve the performance of a variety of natural language processing (NLP) tasks, such as Open IE \cite{Swarnadeep2018,cetto2018graphene}, MT \cite{stajner2016can,stajner2018improvingMT}, Relation Extraction \cite{Miwa2010simplificationRE}, Semantic Role Labeling \cite{Vickrey2008}, Text Summarization \cite{siddharthan2004syntactic,bouayad2009improving}, Question Generation \cite{heilman2010extracting,bernhard2012question}, or Parsing \cite{Chandrasekar:1996:MMT:993268.993361,jonnalagadda2009towards}. 

Linguistic complexity stems from the use of either a difficult vocabulary or sentence structure. Therefore, TS is classified into two categories: \textit{lexical simplification} and \textit{syntactic simplification}.
Through substituting a difficult word or phrase with a more comprehensible synonym, the former primarily addresses a human audience. Most NLP systems, on the contrary, derive greater benefit from syntactic simplification, which focuses on identifying grammatical complexities in a sentence and converting these structures into simpler ones, using a set of text-to-text rewriting operations. \textit{Sentence splitting} plays a major role here: it divides a sentence into several shorter components, with each of them presenting a simpler and more regular structure that is easier to process for downstream applications.


Many different methods for addressing the task of TS have been presented so far. As noted in \newcite{stajner2017leveraging}, data-driven approaches outperform rule-based systems in the area of lexical simplification \cite{glavas2015,Paetzold:2016:ULS:3016387.3016433,nisioi2017exploring,Zhang2017}. 
In contrast, the state-of-the-art syntactic simplification approaches are rule-based \cite{Siddharthan2014,Ferres2016,Saggion:2015:MSI:2775084.2738046}, providing more grammatical output and covering a wider range of syntactic transformation operations, however, at the cost of being very conservative, often to the extent of not making any changes at all. Acknowledging that existing TS corpora \cite{zhu2010monolingual,Coster:2011:SEW:2002736.2002865,Xu2015newsela} are inappropriate for learning to decompose sentences into shorter, syntactically simplified components, as they contain only a small number of split examples, \newcite{Narayan2017} lately compiled the first TS dataset that explicitly addresses the task of sentence splitting. 
Using this corpus, several encoder-decoder models \cite{bahdanau2014neural} are proposed for breaking down a complex source into a set of sentences with a simplified structure. \newcite{aharoni2018split} further explore this idea, augmenting the presented neural models with a copy mechanism \cite{Gu2016,See2017}.

\begin{figure}[ht]
	\centering
  \includegraphics[width=0.5\textwidth]{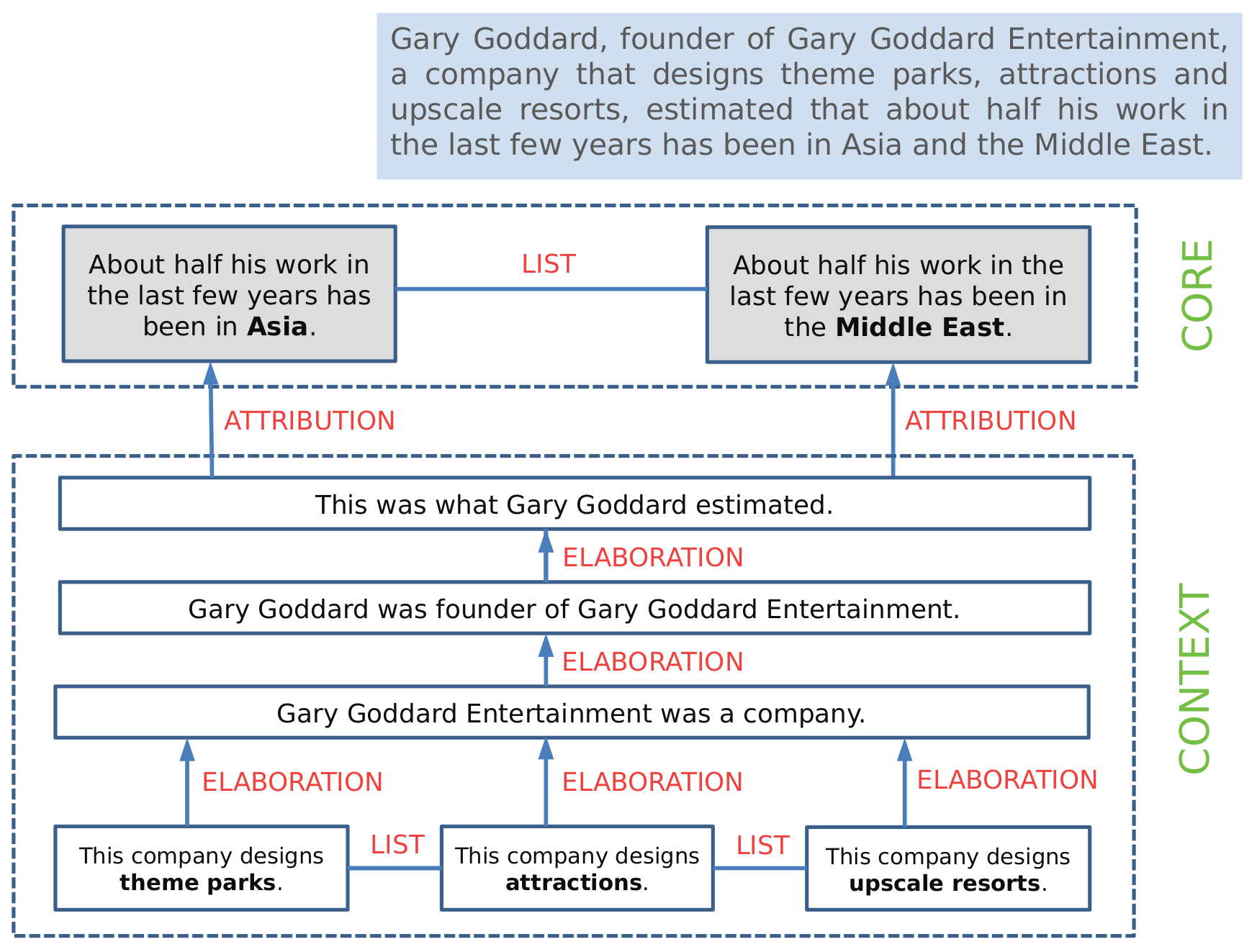}
	\caption{Example of the output that is generated by our proposed TS approach. A complex input sentence is transformed into a semantic hierarchy of simplified sentences in the form of minimal, self-contained propositions that are linked via rhetorical relations.}
	\label{intro_example1}
\end{figure}

In contrast to above-mentioned end-to-end neural approaches, we followed a more systematic approach. First, we performed an in-depth study of the literature on syntactic sentence simplification, followed by a thorough linguistic analysis of the syntactic phenomena that need to be tackled in the sentence splitting task. Next, we materialized our findings into a small set of 35 hand-crafted transformation rules that decompose sentences with a complex linguistic structure into shorter constituents that present a simpler and grammatically sound structure, leveraging downstream semantic applications whose predictive quality deteriorates with sentence length and complexity. 

One of our major goals was to \textit{overcome the conservatism} exhibited by state-of-the-art syntactic TS approaches, i.e. their tendency to retain the input sentence rather than transforming it. For this purpose, we decompose each source sentence into minimal semantic units and turn them into self-contained propositions. In that way, we provide 
a fine-grained output that is easy to process for subsequently applied NLP tools. Another major drawback of the structural TS approaches described so far is that they do not \textit{preserve the semantic links between the individual split components}, resulting in a set of incoherent utterances. Consequently, important contextual information is lost, impeding the interpretability of the output for downstream semantic tasks. To prevent this, we establish a contextual hierarchy between the split components and identify the semantic relationship that holds between them. An example of the resulting output is displayed in Figure \ref{intro_example1}.

\section{Related Work}

To date, three main classes of techniques for syntactic TS with a focus on the task of sentence splitting have been proposed. The first uses a set of syntax-based hand-crafted transformation rules to perform structural simplification operations, while the second exploits machine learning (ML) techniques where the model learns simplification rewrites automatically from examples of aligned complex source and simplified target sentences. In addition, approaches based on the idea of decomposing a sentence into its main semantic constituents using a semantic parser were described.

\subsection{Syntax-driven Rule-based Approaches}
The line of work on structural TS starts with \newcite{Chandrasekar:1996:MMT:993268.993361}, who manually defines a set of rules to detect points where sentences may be split, such as relative pronouns or conjunctions, based on chunking and dependency parse representations. \newcite{siddharthan2002architecture} presents a pipelined architecture for a simplification framework that extracts a variety of clausal and phrasal components from a source sentence and transforms them into stand-alone sentences using a set of hand-written grammar rules based on shallow syntactic features.

More recently, \newcite{Siddharthan2014} propose RegenT, a hybrid TS approach that combines an extensive set of 136 hand-written grammar rules defined over dependency tree structures for tackling 7 types of linguistic constructs with a much larger set of automatically acquired rules for lexical simplification. Taking a similar approach, \newcite{Ferres2016} describe a linguistically-motivated rule-based TS approach called YATS, which relies on part-of-speech tags and syntactic dependency information to simplify a similar set of linguistic constructs, using a set of only 76 hand-crafted transformation patterns in total. 
These two state-of-the-art rule-based structural TS approaches primarily target reader populations with reading difficulties, such as people suffering from dyslexia, aphasia or deafness. According to \newcite{siddharthan2014survey}, 
those groups most notably benefit from splitting long sentences that contain clausal constructions. 
Consequently, simplifying 
clausal components 
is the main focus of the 
proposed TS systems of this category.

Finally, \newcite{stajner2017leveraging} present \textsc{LexEv} and \textsc{EvLex}, which combine a syntactic simplification approach that uses an even smaller set of 11 hand-written rules to perform sentence splitting and deletion of irrelevant sentences or sentence parts with an
unsupervised lexical simplifier based on word embeddings
\cite{glavas2015}.

\subsection{Approaches based on Semantic Parsing}

While the TS approaches described above are based on syntactic information, there are a variety of methods that use semantic structures for sentence splitting. These include the work of \newcite{narayan2014hybrid} and \newcite{Narayan2016}, who propose a framework that 
takes semantically-shared elements as the basis for splitting and rephrasing a sentence. It first generates a semantic representation of the input 
to identify splitting points in the sentence. In a second step, the split components are then rephrased by completing them with missing elements in order to reconstruct grammatically sound sentences. Lately, with DSS, \newcite{sulemSystem} describe another semantic-based structural simplification framework that follows a similar approach. 

\subsection{Data-driven Approaches}
More recently, data-driven approaches for the task of sentence splitting emerged. \newcite{Narayan2017} propose a set of sequence-to-sequence models trained on the WebSplit corpus, a dataset of over one million tuples that map a single complex sentence to a sequence of structurally simplified sentences. 
\newcite{aharoni2018split} further explore this idea, augmenting the presented neural models with a copy mechanism. Though outperforming the models used in \newcite{Narayan2017}, they still perform poorly compared to previous state-of-the-art rule-based syntactic simplification approaches. In addition, \newcite{Botha2018} observed that the sentences from the WebSplit corpus contain fairly unnatural linguistic expressions using only a small vocabulary. To overcome this limitation, they present a scalable, language-agnostic method for mining training data from Wikipedia edit histories, providing a rich and varied vocabulary over naturally expressed sentences and their extracted splits. When training the best-performing model of \newcite{aharoni2018split} on this new split-and-rephrase dataset, they achieve a strong improvement over prior best results from \newcite{aharoni2018split}. However, due to the uniform use of a single split per source sentence in the training set, each input sentence is broken down into two output sentences only.
Consequently, the resulting simplified sentences are still comparatively long and complex.


\section{Recursive Sentence Splitting}


We present \textsc{DisSim}, a recursive sentence splitting approach that creates a semantic hierarchy of simplified sentences.\footnote{The source code of our framework is available under \url{https://github.com/Lambda-3/DiscourseSimplification}.} 
The goal of our approach is to \textbf{generate an intermediate representation that presents a simple and more regular structure which is easier to process for downstream semantic applications and may support a faster generalization in ML tasks}. For this purpose, we \textit{cover a wider range of syntactic constructs} (10 in total) than state-of-the-art rule-based syntactic frameworks. In particular, our approach is not limited to breaking up clausal components, but also splits and rephrases a variety of phrasal elements, resulting in a much more fine-grained output where each proposition represents a minimal semantic unit that is typically composed of a simple subject-predicate-object structure. Though tackling a larger set of linguistic constructs, our framework operates on a \textit{much smaller set of only 35 manually defined rules} as compared to existing syntax-driven rule-based approaches.

With the help of the transformation patterns that we specified, source sentences that present a complex linguistic form are transformed into clean, compact structures by disembedding clausal and phrasal components that contain only supplementary information. These elements are then transformed into independent sentences. In that way, the source sentence is reduced to its key information (\textit{``core sentence''}) and augmented with a number of associated \textit{contextual sentences} that disclose additional information about it, resulting in a novel hierarchical representation in the form of core sentences and accompanying contexts. Moreover, we identify the rhetorical relations by which core sentences and their associated contexts are connected in order to preserve their semantic relationship. The resulting representation of the source text, which we will call a \textit{``discourse tree''} in the following, can then be used to facilitate a variety of artificial intelligence tasks, such as text summarization, MT, IE or opinion mining, among other.



\subsection{Transformation Stage}

The structural TS framework that we propose takes a sentence as input and performs a recursive transformation stage that 
is based upon 35 hand-crafted grammar rules. 
 Each rule defines how to \textit{split up and rephrase} the input into structurally simplified sentences (subtask 1), establish a \textit{contextual hierarchy} between the split components (subtask 2) and identify the \textit{semantic relationship} that holds between those elements (subtask 3). 

\VerbatimFootnotes

The transformation patterns are based on syntactic and lexical features that can be derived from a sentence's phrase structure. They were heuristically determined in a rule engineering process whose main goal was to provide a best-effort set of patterns, targeting the challenge of being applied in a recursive fashion and to overcome biased or incorrectly structured   parse trees. We empirically determined a fixed
execution order of the rules by examining which sequence achieved the best simplification results in a manual qualitative analysis conducted on a development test set of 100 randomly sampled Wikipedia sentences.
The grammar rules are applied recursively in a top-down fashion on the source sentence, until no more simplification pattern matches. In that way, the input is turned into a discourse tree, consisting of \textbf{a set of hierarchically ordered and semantically interconnected sentences that present a simplified syntax}. 
Table~\ref{examplePatterns} displays some examples of our transformation patterns,\footnote{For reproducibility purposes, the complete set of transformation patterns is available under \url{https://github.com/Lambda-3/DiscourseSimplification/tree/master/supplemental_material}.} which are specified in terms of Tregex patterns.\footnote{See \newcite{Levy2006} for details on the rule syntax.} 

\begin{table}[!ht]
\scriptsize
\centering
  \begin{tabular}{| c | l | c | }
    \hline
    & \textsc{Clausal/Phrasal type} & \textsc{\# rules}  \\ \hline \hline
    \multicolumn{3}{|c|} {\textbf{Clausal disembedding}} \\ \hline
    1 & Coordinate clauses & 1 \\ \hline

    2 & Adverbial clauses & 6  \\ \hline
    
    
    
    3a & Relative clauses (non-defining) & 8  \\ \hline
    
    3b & Relative clauses (defining) & 5  \\ \hline
    
    4 & Reported speech & 4  \\ \hline\hline
    
    \multicolumn{3}{|c|} {\textbf{Phrasal disembedding}} \\ \hline 
    5 & Coordinate verb phrases (VPs) & 1 \\ \hline
    6 &  Coordinate noun phrases (NPs) & 2  \\ \hline
    7a &  Appositions (non-restrictive) & 1  \\ \hline
    7b & Appositions (restrictive) & 1  \\ \hline
    
    8 & Prepositional phrases (PPs) & 3  \\ \hline
    9 & Adjectival and adverbial phrases & 2 \\ \hline
    10 & Lead NPs & 1  \\ \hline \hline
    
    
    & Total & 35 \\ \hline

  \end{tabular} 
  
  \caption{Linguistic constructs addressed by \textsc{DisSim}.}
  \label{rulesAndFrequency}
\end{table}

\paragraph{Subtask 1: Sentence Splitting and Rephrasing.}
Each transformation rule takes a sentence's 
phrasal parse tree\footnote{generated by Stanford's pre-trained lexicalized parser \cite{Socher2013}} as input and encodes a pattern that, in case of a match, will extract textual parts from the tree.
The decomposed text spans, as well as the remaining text span are then transformed into new stand-alone sentences.
In order to ensure that the resulting simplified output is grammatically sound, some of the extracted text spans are combined with their corresponding referents from the main sentence or appended to a simple phrase (e.g. \textit{``This is''}). In that way, the simplification rules encode both the splitting points and rephrasing procedure for reconstructing proper sentences. 
Both coordinate and subordinate clauses, as well as various types of phrasal elements are addressed by our TS approach. Table~\ref{rulesAndFrequency} provides an overview of the linguistic constructs that are tackled, including the number of transformation patterns that were specified for the respective syntactic phenomenon. 

\begin{table*}[!ht]
\scriptsize
\centering
  \begin{tabular}{ | p{4.5cm} || p{7cm} | p{3cm} |}
    \hline
      \textsc{Rule} & \textsc{Tregex pattern} & \textsc{Extracted sentence} \\ \hline \hline
    
    \textbf{SharedNPPostCoordinationExtractor} (for coordinate verb phrases) & ROOT $<<:$ (S $<$ (\underline{NP} $\$..$ (VP $<+$(VP) (VP $>$ VP $\$..$ \fbox{VP})))) & \underline{NP} + \fbox{VP}. \\ \hline
    

                       \textbf{SubordinationPreExtractor} (for adverbial clauses with pre-posed subordinative clauses) & ROOT $<<:$ (S $<$ (\textbf{SBAR} $<$ (\fbox{S $<$ (NP $\$..$ VP)}) $\$..$ (NP $\$..$ VP))) & \fbox{S $<$ (NP $\$..$ VP)}. \\  \hline
                       \end{tabular}

  \caption{A selection of transformation rule patterns. A boxed pattern represents the part that is extracted from the input sentence. An underlined pattern designates its referent. A pattern in bold will be deleted from the remaining part of the input.}
  \label{examplePatterns}
\end{table*}

\begin{figure}[!ht]
\centering
\small
\begin{examplebox}{Example: \textsc{SubordinationPreExtractor}}
Input: \inlinetext{Although the Treasury will announce details of the November refunding on Monday, the funding will be delayed if Congress and President Bush fail to increase the Treasury's borrowing capacity.}
\tcbline
Matched Pattern:
\begin{center}
\begin{tikzpicture}[scale=0.6, every tree node/.style={align=center}]
\Tree [.ROOT
        [.S
            [.SBAR
                    [.IN \framebox{Although} ]
                    [.S
                      [.NP \edge[roof]; {the Treasury} ]
                      [.VP \edge[roof]; {will announce \\details ... \\on Monday} ] ] ] 
            [., , ]
            [.NP \edge[roof]; {the funding} ]
            [.VP \edge[roof]; {will be delayed \\if ...\\ borrowing capacity} ]
          [.. . ] ] ]
\end{tikzpicture}
\end{center}
\tcbline
Extraction:
\begin{center}
\begin{tikzpicture}[scale=0.8, level distance=2cm, sibling distance=0.5cm, every tree node/.style={align=center}]
\Tree [.\node[style={draw,rectangle}] {\textbf{(3)} \inlinetext{although} $\rightarrow$ \textit{Contrast}}; 
  \edge node[midway, left] {\textbf{(2)} context}; {\textbf{(1)}\\ The Treasury will announce\\ details of the November\\ refunding on Monday.}
  \edge node[midway, right] {\textbf{(2)} core}; {\textbf{(1)}\\The funding will be delayed if \\ Congress and President Bush fail\\ to increase the  Treasury's\\ borrowing capacity.}
]
\end{tikzpicture}
\end{center}
\end{examplebox}

\caption{\textbf{(Subtask 1)} The source sentence is split up and rephrased into a set of syntactically simplified sentences. \textbf{(Subtask 2)} Then, the split sentences are connected with information about their constituency type to establish a contextual hierarchy between them. \textbf{(Subtask 3)} Finally, by identifying and classifying the rhetorical relations that hold between the simplified sentences, their semantic relationship is restored which can be used to inform downstream applications.}
\label{fig:subordination_post_example}
\end{figure}
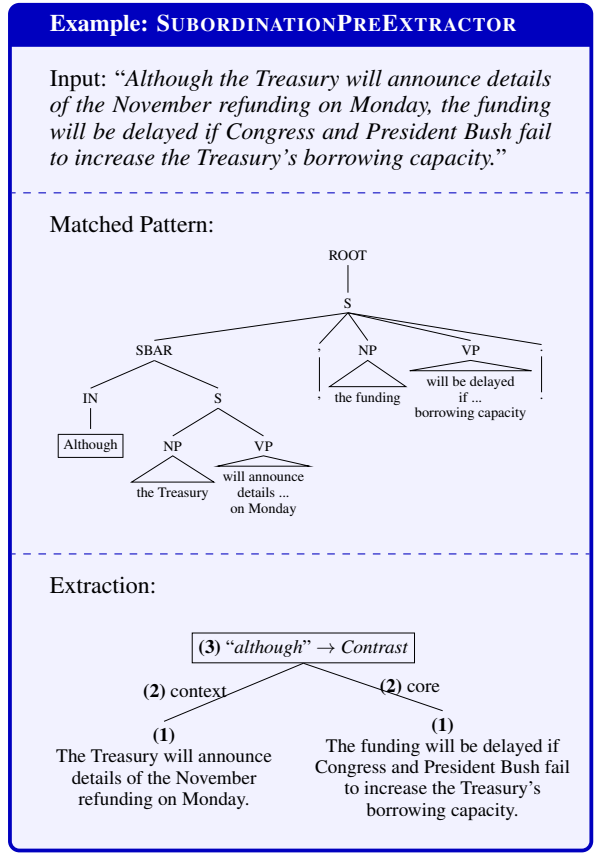

For a better understanding of the splitting and rephrasing procedure, Figure~\ref{fig:subordination_post_example} visualizes the application of the first grammar rule that matches the given input sentence. 
The upper part of the box represents the complex input, which is matched against the simplification pattern. The lower part then depicts the result of the transformation operation.

\paragraph{Subtask 2: Constituency Type Classification.}

Each split will create two or more sentences with a simplified syntax. In order to establish a contextual hierarchy between them, we connect them with information about their constituency type. 
According to \newcite{collinsgrammar}, clauses can be related to one another in two ways: First, there are parallel clauses that are linked by coordinating conjunctions, and second, clauses may be embedded inside another, introduced by subordinating conjunctions. The same applies to phrasal elements. Since the latter commonly express minor information, we denote them \textit{context sentences}. In contrast, the former are of equal status and typically depict the key information contained in the input. Therefore, they are called \textit{core sentences} in our approach. To differentiate between those two types of constituents, the transformation patterns encode a simple syntax-based approach where subordinate clauses and phrasal elements are classified as \textit{context} sentences, while coordinate clauses/phrases are labelled as \textit{core}.\footnote{This approach roughly relates to the concept of nuclearity in Rhetorical Structure Theory (RST) \cite{mann1988rhetorical}, which specifies each text span as either a \textit{nucleus} or a \textit{satellite}. The nucleus span embodies the central piece of information, whereas the role of the satellite is to further specify the nucleus.}




\paragraph{Subtask 3: Rhetorical Relation Identification.}
Finally, we aim to determine intra-sentential semantic relationships in order to restore semantic relations between the disembedded components. For this purpose, we identify and classify the rhetorical relations that hold between the simplified sentences, making use of both syntactic and lexical features which are encoded in the transformation patterns. While syntactic features are manifested in the phrasal composition of a sentence's parse tree, lexical features are extracted from the parse tree in the form of \textit{cue phrases}. The determination of potential cue words and their positions in specific syntactic environments is based on the work of \newcite{knott1994using}. The extracted cue phrases are then used to infer the type of rhetorical relation. For this task we utilize a predefined list of rhetorical cue words adapted from the work of \newcite{Taboada13}, which assigns them to the relation that they most likely trigger.
For example, the transformation rule in Figure \ref{fig:subordination_post_example} specifies that ``although'' is the cue word here, which is mapped to a ``Contrast'' relationship.


\subsection{Final Discourse Tree}

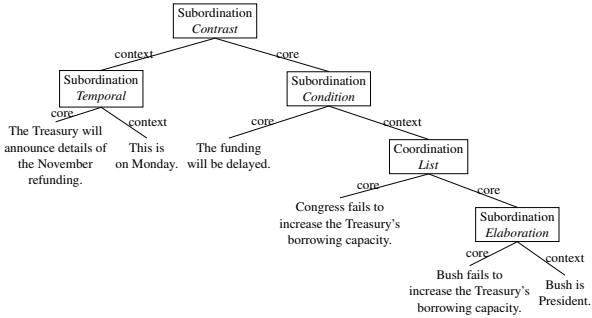
\begin{figure}[!ht]
\centering

\begin{tikzpicture}[scale=0.45, level distance=2cm, sibling distance=0cm, every tree node/.style={align=center, transform shape}]
\Tree [
      .\node [style={draw,rectangle}] {Subordination\\\textit{Contrast}};
            \edge node[midway, left] {context}; [
            .\node [style={draw,rectangle}] {Subordination\\\textit{Temporal}};
                        \edge node[midway, left] {core}; [.\node(d){The Treasury will\\announce details of\\the November\\ refunding.};]
                        \edge node[midway, right] {context}; [.\node(e){This is\\ on Monday.};]
            ]
            \edge node[midway, right] {core}; [
              .\node [style={draw,rectangle}] {Subordination\\\textit{Condition}};
                \edge node[midway, left] {core}; [.\node(b){The funding\\will be delayed.};]
                \edge node[midway, right] {context}; [
                  .\node [style={draw,rectangle}] {Coordination\\\textit{List}};
                    \edge node[midway, left] {core}; [.\node(c){Congress fails to\\increase the Treasury's\\borrowing capacity.};]
                    \edge node[midway, right] {core};[.\node [style={draw,rectangle}] {Subordination\\\textit{Elaboration}};
                        \edge node[midway, left] {core}; [.\node(d){Bush fails to\\increase the Treasury's\\borrowing capacity.};]
                        \edge node[midway, right] {context}; [.\node(e){Bush is\\ President.};]
                        ]
                        ]
                ]
            ]
        ]
]
\end{tikzpicture}
\caption{Final discourse tree of the example sentence.}
\label{example_final}
\end{figure}

The leaf nodes resulting from the first simplification pass are recursively simplified in a top-down approach. When no more transformation rule matches, the algorithm stops. The final discourse tree for the example sentence of Figure~\ref{fig:subordination_post_example} is shown in Figure~\ref{example_final}. 


\section{Experimental Setup}

To compare the performance of our TS approach with state-of-the-art syntactic simplification systems, we evaluated \textsc{DisSim} with respect to the \textit{sentence splitting task} (subtask 1). The evaluation of the rhetorical structures (subtasks 2 and 3) will be subject of future work.

\paragraph{Corpora.} We conducted experiments on three commonly used simplification corpora from two different domains. The first dataset we used was Wikilarge, which consists of 359 sentences from the PWKP corpus \cite{xu2016optimizing}.  Moreover, to demonstrate domain independence, we compared the output generated by our TS approach with that of the various baseline systems on the Newsela corpus \cite{Xu2015newsela}, which is composed of 1077 sentences from newswire articles. In addition, we assessed the performance of our simplification system using the 5000 test sentences from the  WikiSplit benchmark \cite{Botha2018}, which was mined from Wikipedia edit histories.

\paragraph{Baselines.} We compared our \textsc{DisSim} approach against several state-of-the-art baseline systems that have a strong focus on syntactic transformations through explicitly modeling splitting operations. For Wikilarge, these include (i) 
DSS; 
 (ii) SENTS \cite{sulemSystem}, which is an extension of DSS that runs the split sentences through the NTS system \cite{nisioi2017exploring}; (iii) \textsc{Hybrid} \cite{narayan2014hybrid}; (iv)
YATS; and (v) RegenT.
In addition, we report evaluation scores for the complex input sentences, which allows for a better judgment of system conservatism, and the corresponding simple reference sentences.
 With respect to the Newsela dataset, we considered the same baseline systems, with the exceptions of DSS and SENTS, whose outputs were not available. Finally, regarding the WikiSplit corpus, we restricted the comparison to the best-performing system in \newcite{Botha2018}, Copy512, which is a sequence-to-sequence neural model augmented with a copy mechanism and trained over the Wiki\-Split dataset.

\paragraph{Automatic Evaluation.} The automatic metrics that were calculated in the evaluation procedure comprise a number of basic statistics, including (i) the average sentence length of the simplified sentences in terms of the average number of tokens per output sentence (\#T/S); (ii) the average number of simplified output sentences per complex input (\#S/C); (iii) the percentage of sentences that are copied from the source without performing any simplification operation (\%SAME), serving as an indicator for system conservatism; 
and (iv) the averaged Levenshtein distance from the input (LD\textsubscript{SC}), which provides further evidence for a system's conservatism. Furthermore, in accordance with prior work on TS, we report average BLEU \cite{papineni2002bleu} and SARI \cite{xu2016optimizing} scores for the rephrasings of each system.\footnote{For the computation of the BLEU and SARI scores we used the implementation of \newcite{nisioi2017exploring} which is available under \url{https://github.com/senisioi/NeuralTextSimplification}.} 
 Finally, we computed the SAMSA and SAMSA\textsubscript{abl} score of each system, which are the first metrics that explicitly target syntactic aspects of TS \cite{sulemsemantic}.
\paragraph{Manual Analysis.} Human evaluation is carried out on a subset of 50 randomly sampled sentences per corpus by 2 non-native, but fluent English speakers who rated each input-output pair according to three parameters: grammaticality (G), meaning preservation (M) and structural simplicity (S) (see Section \ref{sec:annotation_guidelines} of the appendix). 



In order to get further insights into the quality of our implemented simplification patterns, we performed an extensive \textit{qualitative analysis of the 35 hand-crafted transformation rules}, 
comprising a manual recall-based analysis of the simplification patterns, and a detailed error analysis. 

\paragraph{Usefulness.} Since the \textsc{DisSim} framework that we propose is aimed at serving downstream semantic applications, we measure if an improvement in the performance of NLP tools is achieved when using our TS approach as a preprocessing step. For this purpose, we chose the task of Open IE \cite{Banko07} and determine whether such systems benefit from the sentence splitting approach presented in this work.


\section{Results and Discussion}


\begin{table*}[!ht]
\begin{minipage}[b]{0.67\textwidth}
\scriptsize
\flushleft
  \begin{tabular}{ | p{0.95cm} || c | c | c | c | c | c | c | c | }
    \hline
    & \#T/S & \#S/C & \%\newline SAME & LD\textsubscript{SC} & BLEU & SARI & SAM\-SA & SAM\-SA\textsubscript{abl} \\ \hline \hline
    \multicolumn{9}{|c|}{\textbf{359 test sentences from the Wikilarge corpus}} \\\hline
    \cellcolor{gray!25}{Complex} & \cellcolor{gray!25}{22.06} & \cellcolor{gray!25}{1.03} & \cellcolor{gray!25}{100} & \cellcolor{gray!25}{0.00}  & \cellcolor{gray!25}{94.25} & \cellcolor{gray!25}{32.53} & \cellcolor{gray!25}{0.59} & \cellcolor{gray!25}{0.96}\\ \hline
    \cellcolor{gray!25}{Simple reference} & \cellcolor{gray!25}{20.19} & \cellcolor{gray!25}{1.14} & \cellcolor{gray!25}{0.00} & \cellcolor{gray!25}{7.14}  & \cellcolor{gray!25}{99.48} & \cellcolor{gray!25}{43.09} & \cellcolor{gray!25}{0.48} & \cellcolor{gray!25}{0.78}  \\ \hline 
    \textsc{DisSim} & \textbf{11.01} & \textbf{2.82} & \textbf{0.00} & 11.90           & 63.03          & \textbf{35.05} & \textbf{0.67} & 0.84 \\ \hline
    
    DSS                      & 12.91          & 1.87          & \textbf{0.00} & 8.14             & 74.42          & 34.32 & 0.64          & 0.75\\ \hline
    
    SENTS               & 14.17          & 1.09          & \textbf{0.00} & \textbf{13.79}       & 54.37          & 29.76 & 0.40          & 0.58   \\ \hline 
    
    \textsc{Hybrid}               &    13.44       &    1.03      &  \textbf{0.00} &  13.04      &    48.97      & 26.19  &    0.47     & 0.76   \\ \hline 
    
    YATS                     & 18.83          & 1.40          & 18.66        & 4.44          & 73.07          & 33.03 & 0.56          & 0.80  \\ \hline
    
    RegenT                   & 18.20          & 1.45          & 41.50        & 3.77     & \textbf{82.49}          & 32.41 & 0.61          & \textbf{0.85}\\ \hline\hline
    
   
  

    \multicolumn{9}{|c|}{\textbf{1077 test sentences from the Newsela corpus}} \\\hline
    \cellcolor{gray!25}{Complex} & \cellcolor{gray!25}{23.34} & \cellcolor{gray!25}{1.01} & \cellcolor{gray!25}{100} & \cellcolor{gray!25}{0.00}  & \cellcolor{gray!25}{20.91} & \cellcolor{gray!25}{9.84} & \cellcolor{gray!25}{0.49} & \cellcolor{gray!25}{0.96} \\ \hline
    \cellcolor{gray!25}{Simple reference} & \cellcolor{gray!25}{12.81} & \cellcolor{gray!25}{1.01} & \cellcolor{gray!25}{0.00} & \cellcolor{gray!25}{16.25}  & \cellcolor{gray!25}{100} & \cellcolor{gray!25}{91.13} & \cellcolor{gray!25}{0.25} & \cellcolor{gray!25}{0.46}  \\ \hline
    
    \textsc{DisSim} & \textbf{11.20} & \textbf{2.96} & \textbf{0.00} & 13.00           & 14.54          & \textbf{49.00} & \textbf{0.57} & 0.84 \\ \hline
    
     \textsc{Hybrid}   &   12.49         &     1.02      &     \textbf{0.00}     & \textbf{13.46} & 14.42       &   40.34       & 0.38 &       0.74        \\ \hline 
    
    YATS                     & 18.71          & 1.42         & 16.16       & 5.03      & 17.51          & 36.88 & 0.50          & 0.83  \\ \hline
    
    RegenT                   & 16.74          & 1.61          & 33.33       & 5.03   & \textbf{18.96}         & 32.83 & 0.55         & \textbf{0.85} \\ \hline\hline

   
  

    \multicolumn{9}{|c|}{\textbf{5000 test sentences from the WikiSplit corpus}} \\\hline
    \cellcolor{gray!25}{Complex} & \cellcolor{gray!25}{32.01} & \cellcolor{gray!25}{1.10} & \cellcolor{gray!25}{100} & \cellcolor{gray!25}{0.00}  & \cellcolor{gray!25}{74.28} & \cellcolor{gray!25}{29.91} & \cellcolor{gray!25}{0.37} & \cellcolor{gray!25}{0.95} \\ \hline
    \cellcolor{gray!25}{Simple reference} & \cellcolor{gray!25}{18.14} & \cellcolor{gray!25}{2.08} & \cellcolor{gray!25}{0.00} & \cellcolor{gray!25}{7.48}  & \cellcolor{gray!25}{100} & \cellcolor{gray!25}{94.71} & \cellcolor{gray!25}{0.49} & \cellcolor{gray!25}{0.75}  \\ \hline 
    
    \textsc{DisSim} & \textbf{11.91} & \textbf{4.09} & \textbf{0.76} & \textbf{19.10} &  51.96          & 39.33 & \textbf{0.54} & \textbf{0.84} \\ \hline
    
    Copy512                     & 16.55          & 2.08         & 13.30        & 2.39     & \textbf{76.42}          & \textbf{61.51} & 0.51         & 0.78  \\ \hline
   
  \end{tabular} 
  
  \caption{Automatic evaluation results.}
  \label{resultsAutomaticEval}
\end{minipage}
\begin{minipage}[b]{0.29\textwidth}
\flushright
\scriptsize
\begin{tabular}{ | p{0.95cm} || c | c | c || c |}
    \hline
    & G & M & S & avg. \\ \hline \hline
    \multicolumn{5}{|c|}{\textbf{Wikilarge test set}} \\ \hline
    
    \cellcolor{gray!25}{Simple reference} & \cellcolor{gray!25}{4.70} & \cellcolor{gray!25}{4.56} & \cellcolor{gray!25}{-0.2} & \cellcolor{gray!25}{3.02} \\ \hline 
    
    \textsc{DisSim} & 4.36 & 4.50 & \textbf{1.30} & \textbf{3.39} \\ \hline
    
    DSS & 3.44 & 3.68 & 0.06 & 2.39 \\ \hline
    
    SENTS & 3.48 & 2.70 & -0.18 & 2.00 \\ \hline
    
    \textsc{Hybrid} & 3.16 & 2.60 & 0.86 & 2.21 \\ \hline
    
    YATS & 4.40 & \textbf{4.60} & 0.22 & 3.07 \\ \hline
    
    RegenT & \textbf{4.64} & 4.56 & 0.28 & 3.16 \\ \hline\hline
    

  

    \multicolumn{5}{|c|}{\textbf{Newsela test set}} \\ \hline
    
    \cellcolor{gray!25}{Simple reference} & \cellcolor{gray!25}{4.92} & \cellcolor{gray!25}{2.94} & \cellcolor{gray!25}{0.46} & \cellcolor{gray!25}{2.77} \\ \hline
    
    \textsc{DisSim} & 4.44 & 4.60 & \textbf{1.38} & \textbf{3.47} \\ \hline
    
    \textsc{Hybrid} & 2.97 & 2.35 & 0.93 & 2.08 \\ \hline
    
    YATS & 4.26 & 4.42 & 0.32 & 3.00 \\ \hline
    
    RegenT & \textbf{4.54} & \textbf{4.70} & 0.62 & 3.29 \\ \hline\hline
    

  

    \multicolumn{5}{|c|}{\textbf{WikiSplit test set}} \\ \hline
    
    \cellcolor{gray!25}{Simple reference} & \cellcolor{gray!25}{4.72} & \cellcolor{gray!25}{4.32} & \cellcolor{gray!25}{0.44} & \cellcolor{gray!25}{3.16} \\ \hline
    
    \textsc{DisSim} & 4.36 & 4.36 & \textbf{1.66} & \textbf{3.46} \\ \hline
    
    Copy512 & \textbf{4.72} & \textbf{4.72} & 0.92 & 3.45 \\ \hline

  \end{tabular} 
  
  \caption{Human evaluation ratings on a random sample of 50 sentences from each dataset.} 
  \label{resultsHumanEval}
    
\end{minipage}

 \end{table*}

\paragraph{Automatic Evaluation.} The upper part of Table \ref{resultsAutomaticEval} reports the results that were achieved on the 359 sentences from the Wikilarge corpus, using a set of automatic metrics. Transforming each sentence of the dataset, our \textsc{DisSim} approach reaches the highest splitting rate among the TS systems under consideration, together with \textsc{Hybrid}, DSS and SENTS. With 2.82 split sentences per input on average, our framework outputs by a large margin the highest number of structurally simplified sentences per source. Moreover, consisting of 11.01 tokens on average, the \textsc{DisSim} approach returns the shortest sentences of all systems. 
The relatively high word-based Levenshtein distance of 11.90 confirms previous findings. 

With regard to SARI, our \textsc{DisSim} framework (35.05) again outperforms the baseline systems. 
However, it is among the systems with the lowest BLEU score (63.03).
Though, \newcite{sulemBLEU2018} recently demonstrated that BLEU is inappropriate for the evaluation of TS approaches when sentence splitting is involved, since 
it negatively correlates with structural simplicity, thus penalizing sentences that present a simplified syntax, and presents no correlation with the grammaticality and meaning preservation dimensions. For this reason, we only report these scores for the sake of completeness and to match past work.
According to \newcite{sulemsemantic}, the recently proposed SAMSA and SAMSA\textsubscript{abl} scores are better suited for the evaluation of the sentence splitting task. With a score of 0.67, the \textsc{DisSim} framework shows the best performance for SAMSA, while its score of 0.84 for SAMSA\textsubscript{abl} is just below the one obtained by the RegenT system (0.85).\footnote{According to \newcite{sulemsemantic}, SAMSA highly correlates with human judgments for S and G, while SAMSA\textsubscript{abl} achieves the highest correlation for M.}

The results on the Newsela dataset, depicted in the middle part of Table \ref{resultsAutomaticEval}, support our findings on the Wikilarge corpus, indicating that our TS approach can be applied in a domain independent manner.
The lower part of Table \ref{resultsAutomaticEval} illustrates the numbers achieved on the WikiSplit dataset. Though the Copy512 system beats our approach in terms of BLEU and SARI, the remaining scores are clearly in favour of the \textsc{DisSim} system. 

\paragraph{Manual Analysis.} The results of the human evaluation are displayed in Table~\ref{resultsHumanEval}. 
The inter-annotator agreement was calculated using Cohen's $\kappa$, resulting in rates of 0.72 (G), 0.74 (M) and 0.60 (S). 
The assigned scores demonstrate that our \textsc{DisSim} approach outperforms all other TS systems in the S dimension. With a score of 1.30 on the Wikilarge sample sentences, it is far ahead of the baseline approaches, with \textsc{Hybrid} (0.86) coming closest. However, this system receives the lowest scores for G and M. RegenT obtains the highest score for G (4.64), while YATS is the best-performing approach in terms of M (4.60). However, with a rate of only 0.22, it achieves a low score for S, indicating that the high score in the M dimension is due to the conservative approach taken by YATS, resulting in only a small number of simplification operations. This explanation also holds true for RegenT's high mark for G. Still, our \textsc{DisSim} approach follows closely, with a score of 4.50 for M and 4.36 for G, suggesting that it obtains its goal of returning fine-grained simplified sentences that achieve a high level of grammaticality and preserve the meaning of the input. Considering the average scores of all systems under consideration, our approach is the best-performing system (3.39), followed by RegenT (3.16). The human evaluation ratings on the Newsela and Wiki\-Split sentences show similar results, again supporting the domain independence of our proposed approach.




The results of the recall-based qualitative analysis of the transformation patterns, together with the findings of the error analysis are illustrated in Section \ref{error_analysis} of the appendix in Tables \ref{tab:recall} and \ref{tab:error_analysis}.
Concerning the quality of the implemented simplification rules, the percentage of sentences that were correctly split was approaching 100\% for coordinate and adverbial clauses, and exceeded 80\% on average.

\begin{figure}[htb]
	\centering
	\includegraphics[width=0.5\textwidth]{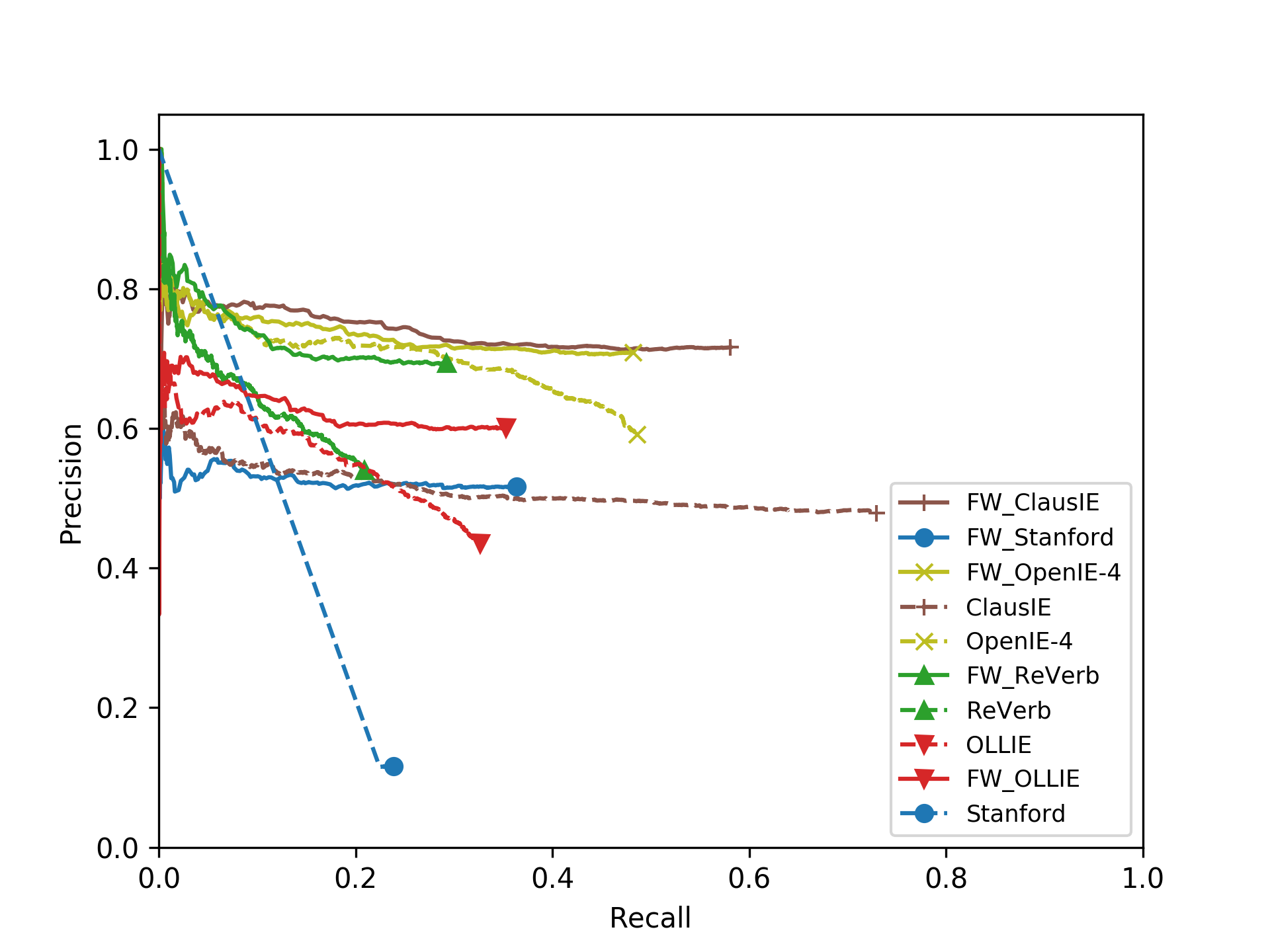}  
  \caption{Performance of state-of-the-art Open IE systems \textit{with} (solid lines) and \textit{without} (dashed lines) sentence splitting as a preprocessing step.}
  \label{fig:OIEBenchmarkSystems}
\end{figure}

\begin{table}[!ht]
\centering
\scriptsize
\begin{tabular}{| l | c | c | c |}
\hline
System & Precision & Recall & AUC \\
\hline
Stanford Open IE           & \textbf{+ 346\%} & \textbf{+ 52\%} & \textbf{+ 597\%} \\
\textsc{ReVerb}            & + 28\%           & + 40\%          & + 57\%           \\
\textsc{Ollie}             & + 38\%           & + 8\%           & + 20\%           \\
ClausIE                    & + 50\%           & - 20\%          & + 15\%           \\
OpenIE-4                   & + 20\%           & - 1\%           & + 3\%            \\
\hline
\end{tabular}
\caption{Improvements when using \textsc{DisSim} as a preprocessing step.}
\label{tab:system_improvements}
\end{table}

\paragraph{Usefulness.} \label{usefulness} To investigate whether our proposed structural TS approach is able to improve the performance of downstream NLP tasks, we compare the performance of a number of state-of-the-art Open IE systems, including ClausIE \cite{DelCorro13}, OpenIE-4 \cite{Mausam16}, \textsc{ReVerb} \cite{Fader11}, \textsc{Ollie} \cite{Mausam12} and Stanford Open IE \cite{Angeli15}, when directly operating on the raw input data with their performance when our \textsc{DisSim} framework is applied as a preprocessing step. 
For this purpose, we made use of the Open IE benchmark framework proposed in \newcite{Stanovsky2016EMNLP}.\footnote{In \newcite{cetto2018graphene}, we further present the performance of our system using the matching function that was originally described in \newcite{Stanovsky2016EMNLP}, which uses a more fine-grained metric for the comparison of relational phrases and arguments.}


The results are displayed in Figure \ref{fig:OIEBenchmarkSystems}. The resulting improvements in overall precision, recall and area under the curve (AUC) are listed in Table \ref{tab:system_improvements}. The numbers show that when using our \textsc{DisSim} framework, all systems under consideration gain in AUC. The highest improvement in AUC was achieved by Stanford Open IE, yielding a 597\% increase over the output produced when acting as a stand-alone system. AUC scores of \textsc{ReVerb} and \textsc{Ollie} improve by 57\% and 20\%. While \textsc{ReVerb} primarily profits from a boost in recall (+40\%), ClausIE, \textsc{Ollie} and OpenIE-4 mainly improve in precision (+50\%, +38\% and +20\%).

\section{Comparative Analysis}
In the following, 
we compare our TS framework with state-of-the-art rule-based syntactic TS approaches and discuss the strengths and weaknesses of each system.

\paragraph{Sentence Splitting.} Table~\ref{outputexampleRuleBased} compares the output generated by the TS systems RegenT and YATS on a sample sentence. As can be seen, RegenT and YATS break down the input into a sequence of sentences that present its message in a way that is easy to digest for human readers. However, the sentences are still rather long and present an irregular structure that mixes multiple semantically unrelated propositions, potentially causing problems for downstream tasks. On the contrary, our fairly aggressive simplification strategy that splits a source sentence into a large set of very short sentences\footnote{In the output generated by \textsc{DisSim}, contextual sentences are linked to their referring sentences and semantically classified by rhetorical relations. The number indicates the sentences' context layer \textit{cl}. Sentences with \textit{cl} = 0 carry the core information of the source, whereas sentences with a \textit{cl}$\geq$1 provide contextual information about a sentence with a context layer of \textit{cl}-1.} is rather inapt for a human audience and may in fact even hinder reading comprehension. Though, we were able to demonstrate that the transformation process we propose can improve the performance of downstream NLP applications. 

 \begin{table}[!ht]
\scriptsize
  \begin{tabular}{ | l | p{5.8cm} |}
    \hline
    \textsc{System} & \textsc{Output} \\ \hline\hline
    Input & \textit{The house was once part of a plantation and it was the home of Josiah Henson, a slave who escaped to Canada in 1830 and wrote the story of his life.} \\ \hline
    RegenT & The house was once part of a plantation. And it was the home of Josiah Henson, a slave. This slave escaped to Canada in 1830 and wrote the story of his life. \\ \hline
   YATS & The house was once part of a plantation. And it was the home of Josiah Henson. Josiah Henson was a slave who escaped to Canada in 1830 and wrote the story of his life.\\ \hline
    \textsc{DisSim} & 
    \begin{minipage}[htb]{\textwidth}
    \begin{itemize}
    \setlength{\topsep}{0pt}
    \setlength{\itemsep}{0pt}
    \setlength{\partopsep}{0pt}
    \setlength{\parsep}{-10pt}
    \setlength{\parskip}{0pt}
    \item[\#1] 0 \textbf{The house was once part of a plantation.}
    \begin{itemize}
    \setlength{\topsep}{0pt}
    \setlength{\itemsep}{0pt}
    \setlength{\partopsep}{0pt}
    \setlength{\parsep}{-10pt}
    \setlength{\parskip}{0pt}
    \item[L:LIST] \#2
    \end{itemize}
    \item[\#2] 0 \textbf{It was the home of Josiah Henson.}
    \begin{itemize}
    \setlength{\topsep}{0pt}
    \setlength{\itemsep}{0pt}
    \setlength{\partopsep}{0pt}
    \setlength{\parsep}{-10pt}
    \setlength{\parskip}{0pt}
    \item[L:ELABORATION] \#3
    \item[L:LIST] \#1
    \end{itemize}
    \item[\#3] 1 \textbf{Josiah Henson was a slave.}
        \begin{itemize}
        \setlength{\topsep}{0pt}
    \setlength{\itemsep}{0pt}
    \setlength{\partopsep}{0pt}
    \setlength{\parsep}{-10pt}
    \setlength{\parskip}{0pt}
    \item[L:ELABORATION] \#4
    \item[L:ELABORATION] \#6
    \end{itemize}
    \item[\#4] 2 \textbf{This slave escaped to Canada.}
    \begin{itemize}
    \setlength{\topsep}{0pt}
    \setlength{\itemsep}{0pt}
    \setlength{\partopsep}{0pt}
    \setlength{\parsep}{-10pt}
    \setlength{\parskip}{0pt}
    \item[L:TEMPORAL] \#5
    \item[L:LIST] \#6
    \end{itemize}
    \item[\#5] 3 \textbf{This was in 1830.}
    \item[\#6] 2 \textbf{This slave wrote the story of his life.}
    \begin{itemize}
    \setlength{\topsep}{0pt}
    \setlength{\itemsep}{0pt}
    \setlength{\partopsep}{0pt}
    \setlength{\parsep}{-10pt}
    \setlength{\parskip}{0pt}
    \item[L:LIST] \#4
    \end{itemize}
    \end{itemize}
    \end{minipage}
    \\ \hline

\end{tabular} 
  
  \caption{Simplification example (from Newsela).}
  \label{outputexampleRuleBased}
\end{table}





\begin{table}[!ht]
\scriptsize
  \begin{tabular}{ | l | p{5.2cm} |}
    \hline
    \textsc{System} & \textsc{Output} \\ \hline\hline
    Input & \textit{``The amabassador's arrival has not been announced and he flew in complete secrecy,'' the official said.} \\ \hline
    \textsc{LexEv, EvLex} & He arrived in complete secrecy.\\ \hline
    \textsc{DisSim} &  
    \begin{minipage}[htb]{0.3\textwidth}
    \begin{itemize}
    \setlength{\topsep}{0pt}
    \setlength{\itemsep}{0pt}
    \setlength{\partopsep}{0pt}
    \setlength{\parsep}{-10pt}
    \setlength{\parskip}{-2pt}

    \item[\#1] 0 \textbf{The ambassador's arrival has not been announced.}
        \begin{itemize}
    \setlength{\topsep}{0pt}
    \setlength{\itemsep}{0pt}
    \setlength{\partopsep}{0pt}
    \setlength{\parsep}{-10pt}
    \setlength{\parskip}{0pt}
            \item[L:LIST] \#2
            \item[L:ATTRIBUTION] \#3
        \end{itemize}
    \item[\#2] 0 \textbf{He flew in complete secrecy.}
    
    \begin{itemize}
    \setlength{\topsep}{0pt}
    \setlength{\itemsep}{0pt}
    \setlength{\partopsep}{0pt}
    \setlength{\parsep}{-10pt}
    \setlength{\parskip}{0pt}
        \item[L:LIST] \#1
        \item[L:ATTRIBUTION] \#3
    \end{itemize}
    \item[\#3] 1 \textbf{This was what the official said.}
    \end{itemize} 
    \end{minipage}
    \\ \hline
    
\end{tabular} 
  
  \caption{Example \cite{stajner2017leveraging}.}
  \label{outputexampleStajner}
\end{table}

\paragraph{Text Coherence.} The vast majority of syntactic simplification approaches do not take into account discourse-level aspects,
producing a disconnected sequence of simplified sentences which results in a loss of cohesion that makes the text harder to interpret \cite{siddharthan2014survey}. 
However, two notable exceptions have to be mentioned. \newcite{siddharthan2006syntactic} was the first to use discourse-aware cues in one of RegenT's predecessor systems, with the goal of generating a coherent output, e.g. by choosing appropriate determiners (\textit{``\underline{This} slave''} in Table~\ref{outputexampleRuleBased}). 
However, as opposed to our approach, where a semantic relationship is established for each output sentence, only a comparatively low number of sentences is linked by such cue words in \newcite{siddharthan2006syntactic}'s framework (and its successors). 
\textsc{EvLex} and \textsc{LexEv} also operate on the discourse level. They are semantically motivated, eliminating irrelevant information from the input by maintaining only those parts of the input that belong to factual event mentions. 
Our approach, on the contrary, aims to preserve the full informational content of a source sentence, as illustrated in Table \ref{outputexampleStajner}. By distinguishing core from contextual information, we are still able to extract only the key information given in the input.

\section{Conclusion}

We presented a recursive sentence splitting approach that transforms structurally complex sentences into a novel hierarchical representation in the form of core sentences and accompanying contexts that are semantically linked by rhetorical relations. In a comparative analysis, we demonstrated that our TS approach achieves the highest scores on all three simplification corpora with regard to SAMSA (0.67, 0.57, 0.54), and comes no later than a close second in terms of SAMSA\textsubscript{abl} (0.84, 0.84, 0.84), 
two recently proposed metrics targeted at automatically measuring the syntactic complexity of sentences.
These findings are supported by the other scores of the automatic evaluation, as well as the manual analysis. 
In addition, the extrinsic evaluation that was carried out based on the task of Open IE verified that downstream semantic applications profit from making use of our proposed structural TS approach as a preprocessing step.
In the future, we plan to investigate the constituency type classification and rhetorical relation identification steps and port this approach to languages other than English.

\bibliography{acl2019}
\bibliographystyle{acl_natbib}

\appendix

\section{Annotation Guidelines for the Manual Evaluation}
\label{sec:annotation_guidelines}
Table \ref{tab:questions_annotators} lists the questions for the human annotation. Since the focus of our work is on structural rather than lexical simplification, we follow the approach taken in \newcite{sulemSystem} in terms of \textsc{Simplicity} and restrict our analysis to the syntactic complexity of the resulting sentences, which is measured on a scale that ranges from -2 to 2 in accordance with \newcite{nisioi2017exploring}, while neglecting the lexical simplicity of the output sentences. Regarding the \textsc{Grammaticality} and \textsc{Meaning preservation} dimensions, we adopted the guidelines from \newcite{stajner2017leveraging}, with some minor deviations to better reflect our goal of simplifying the \textit{structure} of the input sentences, while \textit{retaining their full informational content}.

\begin{table}[!ht]
\footnotesize
\centering
  \begin{tabular}{ | c| p{4cm}  | c |}
    \hline
    \textsc{Param.} & \textsc{Question} & \textsc{Scale} \\ \hline\hline
    G & Is the output fluent and grammatical? & 1 to 5 \\ \hline
    M & Does the output preserve the meaning of the input? & 1 to 5 \\ \hline 
    S & Is the output simpler than the input, ignoring the complexity of the words? & -2 to 2\\ \hline
  \end{tabular} 
  \caption{Questions for the human annotation.}
  \label{tab:questions_annotators}
\end{table}

\section{Qualitative Analysis of the Transformation Patterns and Error Analysis}
\label{error_analysis}

Tables \ref{tab:recall} and \ref{tab:error_analysis} show the results of the recall-based qualitative analysis of the transformation patterns, together with the findings of the error analysis. These analyses were carried out on a dataset which we compiled.\footnote{The dataset is available under \url{https://github.com/Lambda-3/DiscourseSimplification/tree/master/supplemental_material}.} It consists of 100 Wikipedia sentences per syntactic phenomenon tackled by our TS approach. In the construction of this corpus we ensured that the collected sentences exhibit a great syntactic variability to allow for a reliable predication about the coverage and accuracy of the specified simplification rules. 

Note that we do not consider the rules for disembedding adjectival/adverbial phrases and lead NPs, since an examination of the frequency distribution of the syntactic constructs tackled by our approach over the Wikilarge, Newsela and Wiki\-Split test sentences has shown that these types of constructs occur relatively rarely.

\begin{table}[!ht]
\scriptsize
\centering
  \begin{tabular}{ | l || c | c | c |}
    \hline
    & freq. & \%fired & \%correct trans.  \\ \hline \hline
    \multicolumn{4}{|c|} {\textbf{Clausal disembedding}} \\ \hline
    Coordinate clauses & 113 & 93.8\% & 99.1\% \\ \hline

    Adverbial clauses & 113 & 84.1\% & 96.8\% \\ \hline
    
    
    
    Relative clauses (non-def.) & 108 & 88.9\% & 70.8\% \\ \hline
    
    Relative clauses (defining) & 103 & 86.4\% & 75.3\% \\ \hline
    
    Reported speech & 112 & 82.1\% & 75.0\% \\ \hline\hline
    
    \multicolumn{4}{|c|} {\textbf{Phrasal disembedding}} \\ \hline Coordinate VPs & 109 & 85.3\% & 89.2\% \\ \hline
    Coordinate NPs & 115 & 48.7\% & 82.1\% \\ \hline
    Appositions (non-restrictive) & 107 & 86.0\% & 83.7\%\\ \hline
    Appositions (restrictive) & 122 & 87.7\% & 72.0\% \\ \hline
    
    PPs & 163 & 68.1\% & 75.7\%  \\ \hline \hline
    
    Total & 1165 & 81.1\% & 82.0\% \\
     \hline

  \end{tabular} 
  
  \caption{Recall-based qualitative analysis of the transformation rule patterns. This table presents the results of a manual analysis of the performance of the hand-crafted simplification patterns. The first column lists the syntactic phenomena under consideration, the second column indicates its frequency in the dataset, the third column displays the percentage of the grammar fired, and the fourth column reveals the percentage of sentences where the transformation operation results in a correct split.}
  \label{tab:recall}
\end{table}

\begin{table}[H]
\scriptsize
\centering
  \begin{tabular}{ | p{1cm} || c | c | c | c | c | c |}
    \hline
    & Err. 1 & Err. 2 & Err. 3 & Err. 4 & Err. 5 & Err. 6 \\ \hline \hline
    \multicolumn{7}{|c|} {\textbf{Clausal disembedding}} \\ \hline
    Coordinate clauses & 1 & 0 & 0 & 0 & 0 & 0\\ \hline

    Adverbial clauses & 1 & 1 & 0 & 1 & 0 & 0 \\ \hline
    
    
    
    Relative clauses (non-def.)  & 5 & 8 & 0 & 0 & 14 & 1 \\ \hline
    
    Relative clauses (defining)  & 8 & 8 & 2 & 0 & 5 & 1 \\ \hline
    
    Reported speech & 5 & 1 & 13 & 1 & 2 & 1 \\ \hline\hline
    
    \multicolumn{7}{|c|} {\textbf{Phrasal disembedding}} \\ \hline Coordinate VPs & 4 & 3 & 2 & 1 & 0 & 0 \\ \hline
    Coordinate NPs & 3 & 3 & 0 & 3 & 1 & 0\\ \hline
    Appositions (non-restrictive) & 0 & 5 & 3 & 0 & 7 & 0 \\ \hline
    Appositions (restrictive)  & 1 & 21 & 3 & 0 & 0 & 0 \\ \hline
    
    PPs & 3 & 11 & 4 & 6 & 4 & 0 \\ \hline \hline
    
    Total & 31  & 61  & 27  & 12 & 33 & 3 \\
    & (19\%) & (37\%) & (16\%) & (7\%) & (20\%) & (2\%) \\ \hline

  \end{tabular} 
  
  \caption{Error analysis. This table shows the results of the error analysis conducted on the same dataset. Six types of errors were identified (Error 1: additional parts; Error 2: missing parts; Error 3: morphological errors; Error 4: wrong split point; Error 5: wrong referent; Error 6: wrong order of the syntactic elements).}
  \label{tab:error_analysis}
\end{table}

\end{document}